\documentclass[conference]{IEEEtran}
\IEEEoverridecommandlockouts

\usepackage{cite}
\usepackage{amsmath,amssymb,amsfonts}
\usepackage{algorithmic}
\usepackage{graphicx}
\usepackage{textcomp}
\usepackage{xcolor}
\usepackage{booktabs}
\def\BibTeX{{\rm B\kern-.05em{\sc i\kern-.025em b}\kern-.08em
    T\kern-.1667em\lower.7ex\hbox{E}\kern-.125emX}}
\begin{document}

\title{Typhoon ASR Streaming: Steerable Low-Latency Thai Speech Recognition with Real-Time Shallow Fusion}

\author{\IEEEauthorblockN{Warit Sirichotedumrong, Tanawin Samutsin, Shah Faisal Wani, Sittipong Sripaisarnmongkol, Kunat Pipatanakul}
\IEEEauthorblockA{\textit{Typhoon Team, SCB DataX}, Bangkok, Thailand}
}

\maketitle
\begin{abstract}
Open Thai automatic speech recognition (ASR) is dominated by offline,
Whisper-based models that read the whole utterance before transcribing, ruling out
low-latency uses such as live captioning and voice agents. We present a deployable
system for streaming Thai ASR that lets a user steer its vocabulary at decode time,
without retraining. A widely used open Thai model, trained with full context,
collapses when run as a true stream; we restore streaming with a \emph{cache-aware}
encoder, by converting it or adapting a natively streaming one, and add
a shallow-fusion layer that re-ranks candidates inside the streaming decoder with
a GPU $n$-gram language model and phrase boosting. Across two Thai benchmarks and
two model sizes, the streaming models stay usable where the full-context model
fails, cutting character error rate 4.3--4.5$\times$ at a one-second look-ahead
while running faster than real time. Decode-time steering then lifts keyword
recall from 16.6\% to 20.7\% at no accuracy cost and negligible overhead; most of
the gain comes from an $n$-gram over ordinary training transcripts, which resolves
the written form of code-switched words the model hears but spells inconsistently,
with phrase boosting adding targeted control over rare domain terms.
\end{abstract}

\begin{IEEEkeywords}
streaming speech recognition, cache-aware Transducer, shallow fusion, contextual
biasing, Thai ASR.
\end{IEEEkeywords}

\section{Introduction}
\label{sec:intro}
Open Thai automatic speech recognition (ASR) is dominated by offline,
Whisper-based models~\cite{whisper,thonburian,pathumma}. Their popularity stems from large-scale multilingual pretraining and strong accuracy, but they require the entire utterance before transcribing, making them unsuitable for live captioning, voice agents, and meeting transcription that demand low, predictable latency.
Typhoon ASR Real-time~\cite{typhoon-asr}, our previously released model and one of
the most widely used open Thai checkpoints (over 28{,}000 downloads and nearly one
million requests served through OpenTyphoon.ai), broke from this paradigm with a
FastConformer-Transducer that runs about 45$\times$ cheaper than Whisper Large-v3
at comparable accuracy. 

Yet it is still not truly low-latency: its released
checkpoint is trained with full, bidirectional context, so forced into a real
stream its character error rate (CER) rises from 11.0\% to 62.8\%, and at the
lowest latency it stops emitting valid output.

Thai has no word spaces and admits several
written forms per spoken word, including numbers, the repetition marker (\emph{mai
yamok}), and borrowed words, so transcripts must first be normalized to a single
form. Live systems also need vocabulary control for names and domain terms the
acoustic model rarely heard, yet per-customer retraining is impractical. 
Production English ASR systems address this by biasing the decoder at runtime with an $n$-gram language model or a phrase list~\cite{biasing}. However, the open Thai ASR ecosystem is largely built around Whisper-style autoregressive models, leaving no open implementation, training recipe, or systematic study of $n$-gram language models for decoder biasing in Thai. As a result, it remains unclear how effective such approaches are for steering Thai ASR toward domain-specific vocabulary.

We close both gaps with one deployable system; rather than a new fusion algorithm,
we package a known one for streaming Thai and quantify its benefits. Our
contributions are as follows:
\begin{itemize}
    \item We present a model-agnostic framework for steerable streaming Thai
    ASR: cache-aware recognition with decode-time $n$-gram fusion and phrase
    boosting, plus a recipe that converts a full-context
    Typhoon ASR Real-time into a cache-aware streaming model.
    \item We provide insight through a controlled benchmark across two model sizes
    (115M, 0.6B) and two domains, characterizing the accuracy--latency trade-off and
    showing that a general-domain $n$-gram recovers frequent code-switched terms,
    while keyword synthesis and phrase boosting extend control to rare ones, at no
    cost to the real-time factor.
    \item We release the two streaming models, the fusion and streaming code, and
    the Thai evaluation suite.\footnote{\textbf{Project Page:} 	https://warit-s.github.io/typhoon-asr-streaming/ for demo, model checkpoints, and code.}
\end{itemize}

\section{Related Work}
Low-latency ASR systems build on the RNN-Transducer~\cite{graves2012} paired with a
FastConformer encoder~\cite{rekesh2023}, made streamable by cache-aware chunked
attention~\cite{noroozi2024} rather than a full-context Conformer~\cite{gulati2020}.
We build on the cache-aware line and quantify what the full-context variant loses
when forced to stream.

To customize vocabulary without retraining, shallow fusion adds an external
language-model score during decoding~\cite{gulcehre2015, kannan2018} and contextual
biasing boosts a phrase list~\cite{biasing}. We take the GPU $n$-gram fusion and
phrase boosting from the NeMo toolkit~\cite{nemo, ngpu-lm} and run them inside the
cache-aware streaming decoder rather than only in offline decoding.

\section{System Overview}
\begin{figure}[t]
\centering
\includegraphics[width=\columnwidth]{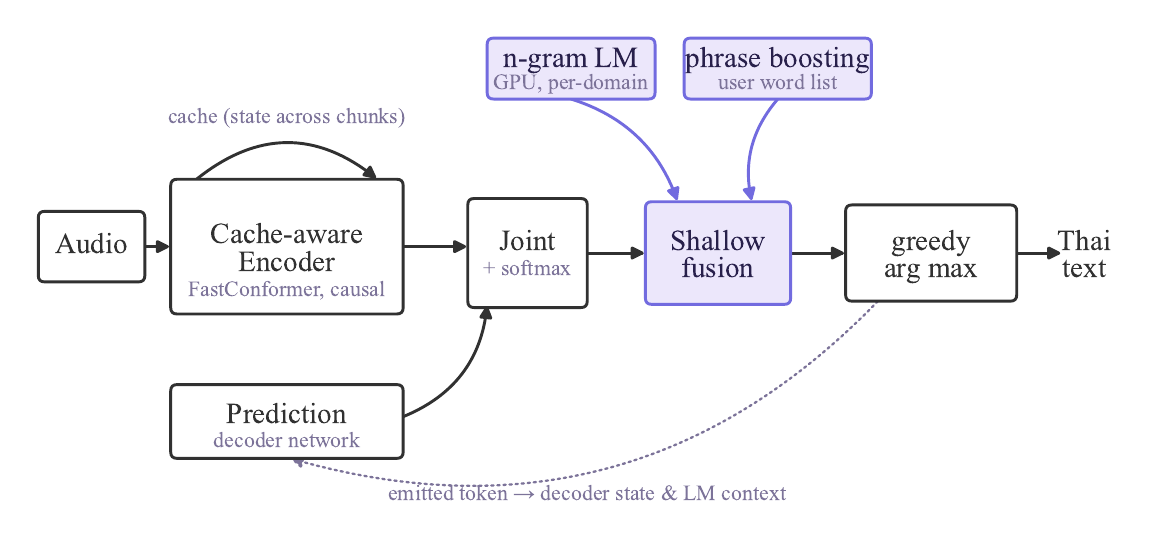}
\caption{System diagram. Audio enters the cache-aware encoder; the prediction
network and joint produce candidate tokens; the $n$-gram language model and the
boosting tree re-rank them at the fusion step; the encoder cache loops across
chunks.}
\label{fig:system_diagram}
\end{figure}

Fig.~\ref{fig:system_diagram} shows the system: audio enters a cache-aware
FastConformer encoder, the prediction and joint networks propose candidates, and a
shallow-fusion layer re-ranks them with a GPU $n$-gram LM and phrase boosting before
greedy decoding emits Thai text. The following subsections cover the data, the two
streaming models, and the fusion layer.

\subsection{Data and Thai Text Normalization}
\label{sec:data}
In Thai, a single spoken word can take several written forms: the repetition
marker (\emph{mai yamok}), numbers, and borrowed words each admit more than
one, so transcripts must be normalized to one canonical form. We resolve this with
the normalization rules from the Typhoon ASR Real-time technical
report~\cite{typhoon-asr}, which map every such case to a single written form. We
train on that report's Thai corpus (Table~\ref{tab:data}): $\sim$11k hours of mostly
large-scale public speech, with smaller curated and synthetic-numeric sets.

\begin{table}[t]
\caption{Thai training corpus from Typhoon ASR Real-time~\cite{typhoon-asr}.}
\begin{center}
\footnotesize
\begin{tabular}{llrr}
\toprule
\textbf{Source} & \textbf{Focus} & \textbf{Hours} & \textbf{Utterances} \\
\midrule
GigaSpeech2~\cite{gigaspeech2}       & Acoustic diversity & 10,329.5 & 9,843,999 \\
Curated media                        & Conversational     & 631.0    & 93,879 \\
Common Voice 17~\cite{commonvoice}   & Read speech        & 35.4     & 31,312 \\
Synthetic TTS                        & Numeric norm.      & 3.2      & 2,697 \\
\midrule
\textbf{Total}                       &                    & \textbf{10,999.1} & \textbf{9,971,887} \\
\bottomrule
\end{tabular}
\label{tab:data}
\end{center}
\end{table}

\subsection{Streaming ASR Models}
\label{sec:streaming}
Both models are FastConformer-Transducer networks~\cite{rekesh2023,gulati2020,graves2012}
in NVIDIA NeMo~\cite{nemo} and decode frame by frame. A full-context encoder attends
over the whole utterance with non-causal convolutions; it is accurate offline but
cannot run on partial audio. A \emph{cache-aware} encoder instead uses chunked,
limited-context attention and causal convolutions, and carries a cache across chunks,
so each chunk costs constant compute~\cite{noroozi2024}.

\textbf{typhoon-asr-streaming-115m.} We continue from Typhoon ASR
Real-time~\cite{typhoon-asr}, our released full-context checkpoint. Converting it to
streaming needs no training: we swap in causal convolutions and chunked,
limited-context attention and copy every weight as a warm start. The converted
model emits almost nothing on its own, so we fine-tune it briefly on the 11k-hour
corpus.

\textbf{typhoon-asr-streaming-nemotron-0.6b.} We adapt NVIDIA's Nemotron streaming ASR
model~\cite{nemotron}, itself a cache-aware FastConformer-Transducer, to Thai. Its
multilingual tokenizer has no Thai sub-words, only characters (consonants, vowels,
and tone marks), so every word is encoded character by character. We therefore \emph{extend} it with the multi-character Thai byte-pair encoding (BPE) from
our Typhoon ASR Real-time tokenizer~\cite{typhoon-asr} rather than replacing it,
which keeps the other languages recoverable, and we add a numeric fine-tuning
stage.

\subsection{Real-Time Shallow Fusion}
Shallow fusion and phrase-list biasing are well-established
ASR techniques~\cite{gulcehre2015, kannan2018, biasing} and are widely used in
production English systems, but remain largely unexplored for open Thai ASR. At each
decoding step we re-rank the decoder's next-token candidates:
\begin{equation}
\mathrm{score}(y) = \log p_{\mathrm{AM}}(y \mid x)
    + \sum_i \alpha_i \log p_{\mathrm{LM},i}(y)
    + \beta \cdot \mathrm{boost}(y),
\end{equation}
where the three terms are the Transducer score, GPU $n$-gram language models
($n=4$, over sub-word tokens, each weighted $\alpha_i$), and a phrase-boosting tree
from a user word list (weight $\beta$). Fusion only re-ranks candidates the model
already proposes.

\section{Experiments}
\label{sec:exp}

\subsection{Setup}
We evaluate on two public Thai test sets from the Typhoon ASR
Benchmark~\cite{typhoon-asr}: \textbf{TVSpeech} (570 utterances of economic and
political Thai speech, rich in names, numbers, and English code-switching) and
\textbf{GigaSpeech2-Thai} (1000 utterances of general YouTube speech). We report
\textbf{CER}, since word error rate is meaningless for space-free
Thai, and, on TVSpeech, \textbf{keyword recall} over the English code-switch
terms in the references, as preliminary experiments showed these
to be the main failure cases. We also report the \textbf{real-time factor (RTF)}, streaming compute
over audio duration, and first-token latency.

All systems run through one streaming evaluation with identical normalization,
carrying the encoder cache across chunks. We sweep the look-ahead from 80 to
3200~ms via the chunk size (one encoder frame is 80~ms).

\textbf{Training.} All training uses NeMo with AdamW, a cosine learning-rate
schedule, and bf16 mixed precision. We fine-tune typhoon-asr-streaming-115m on the
$\sim$11k-hour corpus of Section~\ref{sec:data} for one epoch at a $3{\times}10^{-4}$
peak learning rate and effective batch 128 on four H100s, in about 8.6 hours.
typhoon-asr-streaming-nemotron-0.6b trains in two stages: a one-time Thai adaptation
on the same corpus at a $3{\times}10^{-4}$ peak and effective batch 256 on six H100s,
in about 68 hours, then a short numeric fine-tuning on a 569k-utterance mix at a
$10^{-4}$ peak, in about 3 hours on one H100.

\textbf{Fusion setup.} We set up and tune the fusion as follows:
\begin{enumerate}
\item Build a 66k-sentence in-domain corpus by inserting the 300 English code-switch
terms found in the TVSpeech references into 90 LLM-generated Thai carrier templates;
the same 300 terms form the phrase-boost list.
\item Train two 4-grams over sub-word tokens, neither seeing any test transcript: a
\emph{general} $n$-gram over the ASR training transcripts alone, and a
\emph{keyword-synth} $n$-gram that also includes the in-domain corpus above.
\item Split TVSpeech into development and test halves and select the fusion weights on
development, giving $\alpha{=}\beta{=}0.5$.
\item Apply those weights to the held-out test, evaluating each $n$-gram and the phrase
boost independently against the same no-fusion baseline.
\end{enumerate}

\begin{figure}[t]
\centering
\includegraphics[width=\columnwidth]{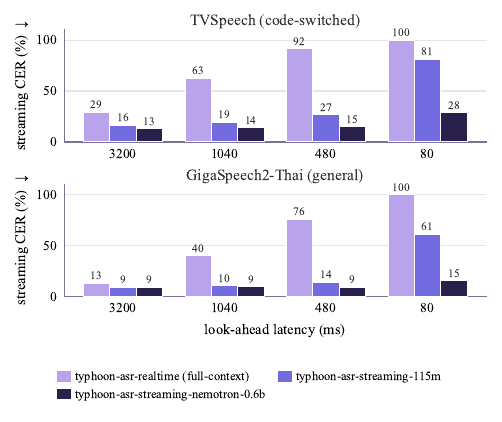}
\caption{Streaming CER (\%) versus look-ahead latency on the full TVSpeech (570
utterances) and GigaSpeech2-Thai (1000 utterances) sets. The full-context model
collapses as latency drops; the streaming models hold. Offline CER (not plotted):
full-context 11.0\% / $\sim$5.8\%; the streaming models sit near their 3200~ms
values.}
\label{fig:cer}
\end{figure}

\subsection{The Full-Context Model Cannot Stream}
Fig.~\ref{fig:cer} shows that streaming the original full-context model is
not viable: its CER degrades from 11.0\% offline to 62.8\% at 1040~ms and 100\% at
80~ms. The streaming models stay usable throughout, and our 0.6B model leads at
every latency, reaching 14.1\% CER at 1040~ms, a 4.5$\times$ reduction over the
streamed baseline. The same collapse-versus-hold pattern holds on general speech
(9.3\% vs.\ 39.8\% at 1040~ms), so the gain is not domain-specific.

\subsection{A General LM Prior Drives the Recall Gain}
As Table~\ref{tab:steering} shows, a domain $n$-gram or phrase-boost list raises
keyword recall on the held-out TVSpeech test from 16.6\% to 20.4\% and 20.7\% at
neutral CER. The steering is also nearly free: adding the $n$-gram and boosting
tree changes the RTF by under 3\% (Table~\ref{tab:rtf}). The gain itself is not circular: a
\emph{general} $n$-gram over ASR transcripts alone, with no keyword synthesis,
already reaches 20.2\% recall, matching the keyword-synth $n$-gram (20.4\%). The
terms it recovers are frequent technology, business, and media words
(\emph{technology}, \emph{subscribe}, \emph{podcast}, \emph{ecosystem}) that the
acoustic model hears but writes inconsistently, as Thai transliterations or variant
spellings (\emph{eco system} for \emph{ecosystem}); having seen them often in the
training transcripts, the $n$-gram supplies the prior that tips the decoder to the
canonical written form. Keyword synthesis therefore matters only for rarer terms
the general transcripts lack, and phrase boosting adds \emph{targeted} per-term
control toward a supplied vocabulary; the upper-bound row (27.8\% recall) marks the
headroom a stronger in-domain LM could still close.

\begin{table}[htbp]
\caption{Decode-time steering on the 0.6B model: TVSpeech \textbf{held-out test}
(285 utterances), 1040~ms.}
\begin{center}
\begin{tabular}{lrr}
\toprule
\textbf{Condition} & \textbf{CER (\%)} $\downarrow$ & \textbf{Recall (\%)} $\uparrow$ \\
\midrule
baseline (no LM)                       & 14.4 & 16.6 \\
+ $n$-gram (general)                   & 14.6 & 20.2 \\
+ $n$-gram (keyword synth)             & 14.6 & 20.4 \\
+ phrase boost                         & \textbf{14.1} & \textbf{20.7} \\
+ $n$-gram on test refs (upper bound)  & 10.2 & 27.8 \\
\bottomrule
\end{tabular}
\label{tab:steering}
\end{center}
\footnotesize \emph{General}: $n$-gram over ASR transcripts only; \emph{keyword synth}
adds a synthetic corpus of the 300 code-switch terms. \emph{Upper bound}: an
$n$-gram trained on the test references, a non-deployable memorization ceiling.
\end{table}

\begin{table}[htbp]
\caption{Efficiency at 1040~ms look-ahead on one NVIDIA H100.}
\begin{center}
\small
\begin{tabular}{lrr}
\toprule
\textbf{Metric} & \textbf{115M} & \textbf{0.6B} \\
\midrule
batch-1 RTF $\downarrow$           & 0.020 & 0.024 \\
\quad{}+ fusion       & 0.020 & 0.024 \\
batch-16 RTF $\downarrow$          & 0.0023 & 0.0031 \\
first-token @1040~ms $\downarrow$  & $\sim$1.06~s & $\sim$1.07~s \\
first-token @480~ms $\downarrow$   & $\sim$0.49~s & $\sim$0.49~s \\
\bottomrule
\end{tabular}
\label{tab:rtf}
\end{center}
\footnotesize \emph{+ fusion}: with $n$-gram fusion and phrase boosting enabled.
Batch-16 RTF is the amortized (throughput) figure; timed over 100 utterances.
\end{table}

\section{Conclusion}
We show that a widely used full-context Thai ASR model collapses when streamed, and
that cache-aware conversion or adaptation restores accuracy across the latency
range; decode-time shallow fusion then adds vocabulary control at negligible cost.
The main insight is on the language-model side: an $n$-gram over ordinary training
transcripts already recovers frequent code-switched terms by pinning inconsistent
spellings to one canonical form, so keyword synthesis and phrase boosting are
needed only for rare, domain-specific vocabulary. Steering only re-ranks
candidates, so it cannot recover sounds the model never produced, and we measure
keyword recall on a single code-switched set. We release the models, code, and
evaluation suite as the low-latency successor to the heavily used full-context
release.

\clearpage
\bibliographystyle{IEEEtran}
\bibliography{references}
\vspace*{\fill}
{\scriptsize © 2026 IEEE. Personal use of this material is permitted. Permission from
IEEE must be obtained for all other uses, in any current or future media,
including reprinting/republishing this material for advertising or
promotional purposes, creating new collective works, for resale or
redistribution to servers or lists, or reuse of any copyrighted component
of this work in other works.}
\end{document}